\pdfoutput=1

\documentclass[11pt]{article}

\usepackage[]{EMNLP2022}

\usepackage{times}
\usepackage{latexsym}
\usepackage{booktabs}
\usepackage{multirow}
\usepackage{tabularx}
\usepackage{xspace}
\usepackage{xcolor,colortbl}
\usepackage{soul}
\usepackage{enumitem}
\usepackage{graphicx}
\usepackage{amsmath}
\usepackage{subcaption}
\usepackage{todonotes}
\usepackage{siunitx}
\usepackage[T1]{fontenc}

\usepackage[utf8]{inputenc}

\usepackage{microtype}

\usepackage{inconsolata}
\definecolor{Gray}{gray}{0.85}
\newcolumntype{g}{>{\columncolor{Gray}}c}

\title{Is (Multilingual) Pretrained Language Model Effectual for Sociodemographic Adaptation?}
\title{Sociodemographic Specialization in the Age of Transformers \\-- Still an Open Issue}
\title{On the Limitation of Sociodemographic Adaptation in the Age of Transformers}
\title{Limitations of Sociodemographic Adaptation in Transformers}
\title{On the Limitations of Sociodemographic Adaptation with Transformers}

\author{Chia-Chien Hung\textsuperscript{1}, Anne Lauscher\textsuperscript{2}, Dirk Hovy\textsuperscript{2}, \\\textbf{Simone Paolo Ponzetto}\textsuperscript{1} and \textbf{Goran Glava\v{s}\textsuperscript{3}} \\
  \textsuperscript{1}Data and Web Science Group, University of Mannheim, Germany \\
  \textsuperscript{2}MilaNLP, Bocconi University, Italy \\
  \textsuperscript{3}CAIDAS, University of Würzburg, Germany \\
  \texttt{\{chia-chien.hung, ponzetto\}@uni-mannheim.de}\\ \texttt{\{anne.lauscher, dirk.hovy\}@unibocconi.it}\\ \texttt{goran.glavas@uni-wuerzburg.de} \\}

\begin{document}
\maketitle
\begin{abstract}
Sociodemographic factors (e.g., gender or age) shape our language.
Previous work showed that incorporating specific sociodemographic factors can consistently improve performance for various NLP tasks in traditional NLP models. We investigate whether these previous  findings still hold with state-of-the-art pretrained Transformers. We use three common specialization methods proven effective for incorporating external knowledge into pretrained Transformers (e.g., domain-specific or geographic knowledge). We adapt the language representations for the sociodemographic dimensions of gender and age, using continuous language modeling and dynamic multi-task learning for adaptation, where we couple language modeling with the prediction of a sociodemographic class.
Our results when employing a multilingual model show substantial performance gains across four languages (English, German, French, and Danish). 
These findings are in line with the results of previous work and hold promise for successful sociodemographic specialization.
However, controlling for confounding factors like domain and language shows that, while sociodemographic adaptation does improve downstream performance, the gains do not always solely stem from sociodemographic knowledge. Our results indicate that sociodemographic specialization, while very important, is still an unresolved problem in NLP.
\end{abstract}


\newcommand{\al}[1]{\textcolor{purple}{#1}}
\section{Introduction}
\label{sec:intro}
Sociodemographic factors like social class, education, income, age, or gender categorize people into specific groups or populations. At the same time, sociodemographic factors both shape and are reflected in our language \cite[e.g.,][]{trudgill, eckert}. For instance, people actively tailor their language to match what is acceptable in groups they identify with (e.g., age groups; cf. \emph{linguistic homophily}). At the same time, many linguistic features are beyond people's control, but mark them clearly as members of a certain group (e.g., dialect terms). 

Various works in NLP have modeled sociodemographic variation, especially the correlations between words and sociodemographic factors~\cite[\emph{inter alia}]{bamman-etal-2014-distributed, garimella-etal-2017-demographic, welch-etal-2020-compositional}. In a similar vein, \newcite{volkova-etal-2013-exploring} and \newcite{hovy-2015-demographic} demonstrated that explicitly modeling demographic factors can consistently improve performance on various tasks. 
However, these observations were made for approaches that leveraged gender-specific lexica to specialize word embeddings and relied on text encoders (e.g., recurrent networks) that have not been pre-trained for language understanding.  
The benefits of demographic specialization have not been tested with Transformer-based~\citep{Transformer} pre-trained language models (PLMs), which have (i)~seen immense text corpora in pre-training and (ii)~been shown to excel on most tasks and even sometimes outperform humans~\citep{wang-etal-2018-glue}.

More recent studies focused mainly on monolingual English datasets and introduce (socio)demographic features in task-specific fine-tuning~\cite{voigt-etal-2018-rtgender, buechel-etal-2018-modeling}, which limits the benefits of (socio)demographic knowledge to tasks at hand. In this work, we investigate the (task-agnostic) sociodemographic specialization of PLMs, aiming to impart the associations between sociodemographic categories and linguistic phenomena into the PLMs parameters: if successful, such specialization would then benefit any downstream NLP task in which sociodemographic factors matter. 
For this, we adopt straight-forward intermediate training paradigms that have been proven effective in specialization of PLMs for other types of knowledge, e.g., in domain, language, and geographic  adaptation~\citep{glavas-etal-2020-xhate, hung-etal-2022-ds,hofmann2022geographic}.

Concretely, 
we encourage the PLMs to establish associations between linguistic phenomena and sociodemographic categories (\textit{gender} and \textit{age} groups, in our case). To this effect, we perform continuous language modeling on specialized corpora and in a dynamic multi-task learning setup~\citep{kendall2018multi}, combining language modeling with the prediction of demographic categories. 

We evaluate the effectiveness of the proposed demographic specialization on three tasks -- we combine demographic category prediction, as an intrinsic evaluation task, with sentiment classification and topic detection as extrinsic evaluation tasks -- and across 4 languages: English, German, French, and Danish. For this, we utilize the multilingual corpus of reviews of \newcite{hovy2015user}, annotated with demographic information. In line with earlier findings \cite{hovy-2015-demographic}, our initial experiments based on a multilingual PLM, multilingual BERT~\citep{devlin-etal-2019-bert}, render demographic specialization effective: we report gains in most tasks and settings. 
Our further analysis shows that, unfortunately, this is just a mirage. Through a set of controlled experiments, in which we (1) adapt with in-domain language modeling alone, without leveraging demographic information and (2) demographically specialize \textit{monolingual} PLMs of evaluation languages, and (3) analyse the topology of the representation spaces of demographically specialized PLMs, we show that most of the original gains can be attributed to confounding effects of language and/or domain specialization.

These findings suggest that specialization approaches proven effective for other types of knowledge fail to adequately instill demographic knowledge into PLMs, making demographic specialization of NLP models an open problem in the age of large pre-trained Transformers.




\section{Sociodemographic Adaptation}
\setlength{\tabcolsep}{3.9pt} 
\label{sec:method}
\begin{table*}[t]
\centering
\footnotesize{
\begin{tabular}{llrrrrrrrr}
\toprule
\multicolumn{1}{l}{} & \multicolumn{1}{l}{} & \multicolumn{4}{c}{\textit{\textbf{gender}}}                                              & \multicolumn{4}{c}{\textit{\textbf{age}}}                                                 \\\cmidrule(lr){3-6}\cmidrule(lr){7-10}
\textbf{Country}     & \textbf{Language}    & \multicolumn{2}{c}{\textbf{Specialization}} & \textbf{SA, AC-SA} & \textbf{TD, AC-TD} & \multicolumn{2}{c}{\textbf{Specialization}} & \textbf{SA, AC-SA} & \textbf{TD, AC-TD} \\ \cmidrule(lr){3-4}\cmidrule(lr){5-6}\cmidrule(lr){7-8}\cmidrule(lr){9-10}
                     &                      & \multicolumn{1}{c}{\textbf{F}}         & \multicolumn{1}{c}{\textbf{M}}        & \multicolumn{2}{c}{\textbf{F / M / X}}      & \multicolumn{1}{c}{\textbf{<35}}         & \multicolumn{1}{c}{\textbf{>45}}        & \multicolumn{2}{c}{\textbf{<35 / >45 / X}}     \\ \midrule
Denmark              & Danish               & 1,596,816            & 2,022,349           & 250,485             & 120,805             & 833,657             & 494,905            & 75,300              & 44,815              \\
France               & French               & 489,778             & 614,495            & 67,305              & 55,570              & 40,448              & 36,182             & 6,570               & 6,120               \\
Germany              & German               & 210,718             & 284,399            & 28,920              & 30,580              & 66,342              & 47,308             & 5,865               & 8,040               \\
UK                   & English              & 1,665,167            & 1,632,894           & 156,630             & 183,995             & 231,905             & 274,528            & 26,325              & 22,095              \\
US                   & English              & 575,951             & 778,877            & 72,270              & 61,585              & 124,924             & 70,015             & 6,495               & 12,090  \\    \bottomrule       
\end{tabular}%
}
\caption{Statistics for the Trustpilot dataset~\citep{hovy2015user} we use in our experiments. For each country (Denmark, France, Germany, UK, and US), we report the language, the size of the specialization portions, and the total number of fine-tuning review texts for each task (sentiment analysis (SA), topic detection (TD), as well as attribute classification on the SA and TD portions (AC-SA and AC-TD, respectively)) for \textit{gender} and \textit{age}. Following~\citet{hovy-2015-demographic}, we split the fine-tuning data randomly into train/dev/test portions with the ratio: 60/20/20.}
\vspace{-0.3em}
\label{tab:dataset_info}
\end{table*}

Our goal is to inject sociodemographic knowledge through intermediate model training in a task-agnostic manner. To achieve this goal, we train the PLM (1) via Masked Language Modeling and (2) in a dynamic multi-task learning setup~\citep{kendall2018multi} where we couple language modeling with the prediction of a sociodemographic factor. This setup pushes the PLM to learn an association (i.e., joint representation) of the contextual information and the sociodemographic factors.

\subsection{Masked Language Modeling (MLM)}
\label{ss:mlm}
Following successful work on pretraining via language modeling for domain-adaptation~\citep{gururangan-etal-2020-dont, hung-etal-2022-ds} 
, we investigate the effect of running standard MLM on the 
data. 
The MLM loss $L_{mlm}$ is computed as the negative log-likelihood of the true token probability:
\vspace{-0.2em}

\small{
\begin{equation}
    L_{mlm}=-\sum_{m=1}^{M}\log P(t_m)\,,
\end{equation}}

\normalsize
\noindent where $M$ is the total number of masked tokens of the given text and $P(t_m)$ is the predicted probability of the token $t_m$ over the vocabulary size. 

\subsection{Sociodemographic Factor Prediction}
\label{ss:socio_predict}
In the multi-task learning setup, the representation of the input texts is additionally fed into a classification head that predicts the sub-group of sociodemographic factors: \textit{age} (under 35 and over 45 -- to avoid fuzzy age boundaries following the setup from \citet{hovy-2015-demographic}), and  \textit{gender} (female and male). The sociodemographic prediction loss $L_{socio}$ is computed as the binary cross-entropy loss:
\vspace{-0.2em}

\small{
\begin{equation}
    L_{socio}=-\sum_{c\in{C}}y(c)\log P(c)\,,
\end{equation}}

\normalsize
\noindent where for each class $c$, we use its predicted probability $P(c)$, and the true class probability $y(c)$.

We investigate the effectiveness of the input text representation for two variants: (i)~using the \texttt{[CLS]} token representation and (ii)~using the contextualized representation of the masked tokens. The \texttt{[CLS]} token represents the whole sequence, while the contextualized representation of the masked tokens focuses on the token-level relationships. Thus, in the first setup, we seek to strengthen the sociodemographic link for a higher-level, wider context, while the token-level setup intuitively enforces a  narrower, lower-level specialization.

\subsection{Multi-Task Learning}
\label{ss:socio_predict}
The multi-task objective forces the model to recognize the sociodemographic aspect while obtaining the linguistic knowledge via masked language modeling. 
In joint multi-task training, we could simply sum the task-specific losses as a na\"ive way of computing a joint loss. However, the summation approach is based on the assumption that all tasks have equal weights of losses (i.e., an equal level of confidence). This does not hold when the losses' scales diverge across different tasks.
Thus, we leverage a dynamic multi-task learning scenario based on the inherent, i.e., \emph{homoscedastic}, uncertainty of the tasks, in which each of the task weights is adjusted automatically to balance the weights~\citep{kendall2018multi}. The intuition is  that the performance can be improved by dynamically assigning less weight to more uncertain tasks, as opposed to assigning uniform task weights throughout the whole training process. This approach  has been proved effective in various NLP tasks~\citep{lauscher-etal-2018-investigating, hofmann2022geographic}. 

In our scenario, $L_{mlm}$ and $L_{socio}$ are measured on different scales in which the model would favor (i.e., be more confident for) one objective than the other. Therefore, the confidence level of the model prediction for each task would change during training progress. 
We dynamically prioritize the tasks by monitoring homoscedastic uncertainties $\sigma_t$: 
\vspace{-0.2em}

\small{
\begin{equation}
\tilde{L}_t = \frac{1}{2\sigma_t^{2}}L_t + \log \sigma_t\,,
\end{equation}}

\normalsize
\noindent where $\sigma_t^2$ is the variance of the task-specific loss over training instances for quantifying the uncertainty of the task $t\in\{mlm, socio\}$. In practice, we train the network to predict the log variance, $\eta_t :=\log \sigma_t^2$, since it is more numerically stable than regressing the variance $\sigma_t^2$ as the log avoids divisions by zero. The losses are then defined as:
\vspace{-0.2em}

\small{
\begin{equation}
\tilde{L}_t = \frac{1}{2}(e^{-\eta_t}L_t + \eta_t)\,.
\end{equation}}

\normalsize
\noindent The final loss is computed as the simple  sum of the two task-specific uncertainty-weighted losses.
\vspace{-0.2em}

\section{Experimental Setup}
\label{s:setup}
\label{sec:experiments}
We describe our setup for assessing the effectiveness of the approaches for intermediate injection of sociodemographic knowledge.

\subsection{Evaluation Tasks and Measure}
We evaluate the effect of the sociodemographic specialization across the three text-classification tasks from~\citet{hovy2015user}. Two of these tasks focus on extrinsic classification and represent a downstream scenario: \textbf{Sentiment Analysis (SA)} and \textbf{Topic Detection (TD)}. Additionally, \textbf{Attribute Classification (AC)} is designed to directly, i.e., intrinsically, evaluate knowledge about sociodemographics by asking the models to predict the sociodemographic class of an author of a text. 

\vspace{0.7em}
\noindent\textbf{Sentiment Analysis (SA)}
is the task of determining the polarity of a given text. We consider SA in product reviews, in which the dataset label is collected based on the 1-, 3-, or 5-star ratings from the review text, which corresponds to a \emph{negative}, \emph{neutral}, and \emph{positive} sentiment, respectively. 

\vspace{0.7em}
\noindent\textbf{Topic Detection (TD)}
is the task of assigning the topic to a given text. We frame TD as a multi-class classification problem on review texts, as before for SA. We consider $5$ classes based on the labels provided by data set we use.

\vspace{0.7em}
\noindent\textbf{Attribute Classification (AC)}
is the intrinsic classification task of identifying the sociodemographic sub-groups based on linguistic features of a text~\citep[e.g., the text is written by a \emph{man} or \emph{woman};][]{ciot-etal-2013-gender, preotiuc-pietro-etal-2015-analysis}. We consider a binary classification formulation given two sociodemographic classes (i.e., predict the correct gender/age sub-group). 

\vspace{0.3em}
\noindent We report the F1-measure for each task.
\setlength{\tabcolsep}{2.3pt} 
\begin{table*}[t]
\centering
\footnotesize{
\begin{tabular}{llcccccgcccccg}
\toprule
                                 &                              & \multicolumn{6}{c}{\textbf{AC-SA}}                                                                                                      & \multicolumn{6}{c}{\textbf{AC-TD}}                                                                                                      \\ \cmidrule(lr){3-8}\cmidrule(lr){9-14}
       \textbf{Model}                          &                              & \textbf{US}          & \textbf{Denmark}     & \textbf{Germany}     & \textbf{France}      & \textbf{UK}          & \textbf{Avg.}        & \textbf{US}          & \textbf{Denmark}     & \textbf{Germany}     & \textbf{France}      & \textbf{UK}          & \textbf{Avg.}        \\\midrule
\multirow{6}{*}{\textit{\textbf{gender}}} & \textit{w/o specialization}  & \multicolumn{1}{l}{} & \multicolumn{1}{l}{} & \multicolumn{1}{l}{} & \multicolumn{1}{l}{} & \multicolumn{1}{l}{} & \multicolumn{1}{l}{} & \multicolumn{1}{l}{} & \multicolumn{1}{l}{} & \multicolumn{1}{l}{} & \multicolumn{1}{l}{} & \multicolumn{1}{l}{} & \multicolumn{1}{l}{} \\
                                 & mBERT                        & 62.6                 & 64.0                 & 59.5                 & 63.9                 & 61.9                 & 62.4                 & 58.1                 & 61.8                 & 57.9                 & 61.2                 & 63.1                 & 60.4                 \\\cmidrule{2-14}
                                 & \textit{with specialization} &                      &                      &                      &                      &                      &         \multicolumn{1}{c}{}              &                      &                      &                      &                      &                      &      \multicolumn{1}{c}{}                 \\
                                 & MLM                          & 63.3                 & \textbf{65.2}                 & 61.2                 & 64.6                 & 63.0                 & 63.4                 & \textbf{59.6}                 & 63.4                 & 60.1                 & 62.1                 & 65.3                 & 62.1                 \\
                                 & MTL-W (CLS)                  & \textbf{63.8}                 & 64.9                 & 60.1                 & 64.1                 & 63.4                 & 63.3                 & 59.2                 & 63.5                 & \textbf{60.3}                 & \textbf{63.1}                 & 64.9                 & \textbf{62.2}                 \\
                                 & MTL-W (CTX)                  & 62.2                 & 65.0                 & \textbf{62.9}                 & \textbf{65.0}                 & \textbf{63.5}                 & \textbf{63.7}                 & 58.8                 & \textbf{63.5}                 & 58.3                 & 62.9                 & \textbf{65.6}                 & 61.8                 \\
                                 \midrule
\multirow{6}{*}{\textit{\textbf{age}}}    & \textit{w/o specialization}  &                      &                      &                      &                      &                      &      \multicolumn{1}{c}{}                 &                      &                      &                      &                      &                      &        \multicolumn{1}{c}{}               \\
                                 & mBERT                        & 62.9                 & 57.2                 & 58.0                 & 55.7                 & 65.1                 & 59.8                 & 60.7                 & 64.5                 & 56.9                 & 56.6                 & 65.2                 & 60.8                 \\\cmidrule{2-14}
                                 & \textit{with specialization} &                      &                      &                      &                      &                      &        \multicolumn{1}{c}{}               &                      &                      &                      &                      &                      &        \multicolumn{1}{c}{}               \\
                                 & MLM                          & \textbf{63.6}                 & \textbf{65.5}                 & \textbf{61.1}                 & \textbf{56.8}                 & \textbf{65.4}                 & \textbf{62.5}                 & \textbf{61.9}                 & 65.1                 & \textbf{58.9}                 & \textbf{57.2}                 & \textbf{65.6}                 & \textbf{61.7}                 \\
                                 & MTL-W (CLS)                  & 60.7                 & 65.2                 & 56.4                 & 55.1                 & 65.3                 & 60.5                 & 61.5                 & \textbf{65.2}                 & 58.2                 & 55.5                 & 62.8                 & 60.6                 \\
                                 & MTL-W (CTX)                  & 59.7                 & 65.3                 & 56.6                 & 54.4                 & 64.0                 & 60.0                 & 61.2                 & 64.6                 & 57.4                 & 55.9                 & 62.8                 & 60.4     \\
                                 \bottomrule
\end{tabular}%
}
\caption{Intrinsic evaluation results for \textit{gender} and \textit{age} on Attribute Classification (AC-SA, AC-TD). We report the F1-score and compare the specialized models (MLM, MTL-W (CLS), MTL-W (CTX)) with vanilla mBERT.}
\label{tab:gender_age_intrinsic}
\vspace{-0.7em}
\end{table*}

\subsection{Data}
We use the dataset introduced by~\citet{hovy2015user} obtained from the  international user review website  Trustpilot.\footnote{\url{https://www.trustpilot.com/}} For investigating the effectiveness of our  sociodemographic specialization methods, we follow the preprocessing and selection criteria from~\citet{hovy-2015-demographic}. This process results in two collections of text for \emph{gender} and \emph{age}, with two sub-groups each: \emph{female} vs. \emph{male}, and \emph{under 35} vs. \emph{over 45},\footnote{As suggested by \citet{hovy-2015-demographic} the split for the age ranges result in roughly equally-sized data sets for each sub-group and is non-contiguous, avoiding fuzzy boundaries.} across 5 countries: the United States (US), Denmark, Germany, France, and United Kingdom (UK).  Thus, our data covers 4 different languages (English, Danish, German, and French).  

To avoid any information leak, we split the collection into task-specific portions~(for SA, TD, and AC, respectively), and portions from which we sample data for model specialization (Specialization). For TD, we avoid topic bias based on sociodemographic groups and eliminate the topic frequency as a confounding factor by restricting the data to the \emph{5 most frequent topics} that occur across the sub-groups for each country. For AC, the task-specific texts correspond to the data used for SA and TD, where the class labels are balanced to minimize the effect of any confounding factors (i.e., we have two AC portions for each sociodemographic factor: \textbf{AC-SA} and \textbf{AC-TD}).  Table~\ref{tab:dataset_info} shows the final number of review texts per country, sociodemographic aspect, and portion in the dataset. 


For intermediate specialization of the multilingual model, we randomly sample 100K instances per sociodemographic group from the \textit{gender} specialization portion and 50K instances each from the texts reserved for \textit{age} specialization concatenated across all 5 countries. For the specialization of monolingual PLMs, we randomly sample 
the same number of instances but from the specialization portions of a \textit{single} country. Following the established procedure~\citep[e.g.,][]{devlin-etal-2019-bert}, we dynamically mask 15\% of the tokens in the sociodemographic specialization portions for MLM.

\setlength{\tabcolsep}{3pt} 
\begin{table*}[t]
\centering
\footnotesize{
\begin{tabular}{llcccccccccccccccg}
\toprule
                                      &                              & \multicolumn{3}{c}{\textbf{US}}               & \multicolumn{3}{c}{\textbf{Denmark}}          & \multicolumn{3}{c}{\textbf{Germany}}          & \multicolumn{3}{c}{\textbf{France}}           & \multicolumn{3}{c}{\textbf{UK}}               &\multicolumn{1}{l}{} \\\cmidrule(lr){3-5}\cmidrule(lr){6-8}\cmidrule(lr){9-11}\cmidrule(lr){12-14}\cmidrule(lr){15-17}
                                       \textit{\textbf{gender}} &  \textbf{Model}                            & \textbf{F}    & \textbf{M}    & \textbf{X} \textbf{}   & \textbf{F}    & \textbf{M}    & \textbf{X}    & \textbf{F}    & \textbf{M}    & \textbf{X}    & \textbf{F}    & \textbf{M}    & \textbf{X}    & \textbf{F}    & \textbf{M}    & \textbf{X}    &  \textbf{Avg.}    \\ \cmidrule{1-17}
\multirow{6}{*}{\textbf{SA}} & \textit{w/o specialization}  &               &               &               &               &               &               &               &               &               &               &               &               &               &               &               & \multicolumn{1}{l}{}              \\
                                      & mBERT                        & 66.3          & 64.4          & 66.0          & 69.2          & 64.8          & 67.2          & 66.1          & 63.2          & 64.5          & 69.3          & 67.0          & 67.8          & 71.0          & 69.0          & 69.7          & 67.0          \\\cmidrule{2-17}
                                      & \textit{with specialization} &               &               &               &               &               &               &               &               &               &               &               &               &               &               &               &   \multicolumn{1}{l}{}            \\
                                      & MLM                          & 67.3          & 66.2          & 66.9          & 69.5          & \textbf{65.8} & 67.8          & \textbf{67.7} & \textbf{65.3} & 66.1          & 69.9          & 67.1          & 68.4          & 72.0          & 70.4          & 71.0          & 68.1          \\
                                      & MTL-W (CLS)                  & 67.2          & 66.3          & 67.0          & \textbf{69.9} & 65.7          & 67.7          & 66.7          & 64.0          & 65.7          & \textbf{70.6} & 67.3          & 68.4          & 72.9          & 70.9          & 71.7          & 68.1          \\
                                      & MTL-W (CTX)                  & \textbf{68.0} & \textbf{66.4} & \textbf{67.3} & 69.1          & 65.6          & \textbf{68.0} & 66.8          & 64.3          & \textbf{66.8} & 70.1          & \textbf{67.5} & \textbf{68.8} & \textbf{73.0} & \textbf{71.0} & \textbf{71.9} & \textbf{68.3} \\ \cmidrule{1-17}
\multirow{6}{*}{\textbf{TD}} & \textit{w/o specialization}  &               &               &               &               &               &               &               &               &               &               &               &               &               &               &               & \multicolumn{1}{l}{}              \\
                                      & mBERT                        & 71.2          & 68.4          & 70.2          & 59.3          & 58.3          & 59.0          & 67.8          & 65.6          & 65.8          & 44.6          & 42.4          & 43.1          & 70.4          & 67.9          & 68.9          & 61.5          \\\cmidrule{2-17}
                                      & \textit{with specialization} &               &               &               &               &               &               &               &               &               &               &               &               &               &               &               & \multicolumn{1}{l}{}              \\
                                      & MLM                          & 72.1          & 69.4          & 70.3          & 59.7          & 58.8          & \textbf{59.4} & \textbf{68.6} & 67.0          & \textbf{67.1} & 45.8          & 43.3          & 44.3          & 70.6          & 67.9          & 69.8          & 62.3          \\
                                      & MTL-W (CLS)                  & 72.3          & 69.2          & 70.4          & 59.7          & 57.8          & 59.1          & 67.6          & 65.7          & 66.4          & \textbf{46.0} & 43.4          & 44.2          & 70.6          & 68.2          & 69.8          & 62.0          \\
                                      & MTL-W (CTX)                  & \textbf{72.8} & \textbf{69.5} & \textbf{70.5} & \textbf{59.9} & \textbf{58.9} & 59.0          & 68.3          & \textbf{67.0} & 66.7          & 45.5          & \textbf{43.9} & \textbf{44.4} & \textbf{70.8} & \textbf{68.2} & \textbf{69.9} & \textbf{62.4}\\
                                      \bottomrule
\end{tabular}%
}
\caption{Results of the extrinsic evaluation on \textit{gender} data for gender-specialized and non-specilized models for Sentiment Analysis (SA) and Topic Detection (TD). We report the F1-score.}
\vspace{-0.3em}
\label{tab:gender_task}
\end{table*}

\setlength{\tabcolsep}{3.4pt} 
\begin{table*}[t]
\centering
\footnotesize{
\begin{tabular}{llcccccccccccccccg}
\toprule
                                   &                              & \multicolumn{3}{c}{\textbf{US}}                                    & \multicolumn{3}{c}{\textbf{Denmark}}                               & \multicolumn{3}{c}{\textbf{Germany}}                               & \multicolumn{3}{c}{\textbf{France}}                                & \multicolumn{3}{c}{\textbf{UK}}                        &   \multicolumn{1}{l}{}      \\\cmidrule(lr){3-5}\cmidrule(lr){6-8}\cmidrule(lr){9-11}\cmidrule(lr){12-14}\cmidrule(lr){15-17}
                                   \textit{\textbf{age}} &       \textbf{Model}                       & \textbf{<35}           & \textbf{>45}           & \textbf{X}           & \textbf{<35}           & \textbf{>45}           & \textbf{X}           & \textbf{<35}           & \textbf{>45}           & \textbf{X}           & \textbf{<35}           & \textbf{>45}           & \textbf{X}           & \textbf{<35}           & \textbf{>45}           & \textbf{X}           & \textbf{Avg.}           \\\cmidrule{1-17}
\multirow{6}{*}{\textbf{SA}} & \textit{w/o specialization}  & \multicolumn{1}{l}{} & \multicolumn{1}{l}{} & \multicolumn{1}{l}{} & \multicolumn{1}{l}{} & \multicolumn{1}{l}{} & \multicolumn{1}{l}{} & \multicolumn{1}{l}{} & \multicolumn{1}{l}{} & \multicolumn{1}{l}{} & \multicolumn{1}{l}{} & \multicolumn{1}{l}{} & \multicolumn{1}{l}{} & \multicolumn{1}{l}{} & \multicolumn{1}{l}{} & \multicolumn{1}{l}{} & \multicolumn{1}{l}{} \\
                                   & mBERT                        & 57.7                 & 57.9                 & 57.8                 & 62.7                 & 62.7                 & 62.9                 & 52.6                 & 55.0                 & 55.0                 & 59.6                 & 57.4                 & 61.5                 & 63.8                 & 63.9                 & 63.7                 & 59.6                 \\\cmidrule{2-17}
                                   & \textit{with specialization} &                      &                      &                      &                      &                      &                      &                      &                      &                      &                      &                      &                      &                      &                      &                      &        \multicolumn{1}{l}{}              \\
                                   & MLM                          & 59.4                 & 57.8                 & \textbf{58.2}        & 63.3                 & 62.1                 & 63.0                 & 53.6                 & 55.5                 & 56.7                 & 59.9                 & 59.5                 & 61.6                 & 62.8                 & 62.0                 & 63.0                 & 59.9                 \\
                                   & MTL-W (CLS)                  & 59.3                 & 57.9                 & 58.0                 & 63.1                 & 62.9                 & 63.0                 & \textbf{53.8}        & 55.3                 & 55.5                 & 60.4                 & \textbf{60.3}        & \textbf{62.8}        & 63.8                 & 64.9                 & 64.9                 & 60.4                 \\
                                   & MTL-W (CTX)                  & \textbf{59.9}        & \textbf{58.6}        & 57.8                 & \textbf{64.2}        & \textbf{63.3}        & \textbf{63.2}        & 53.0                 & \textbf{56.5}        & \textbf{56.7}        & \textbf{60.9}        & 59.8                 & 59.7                 & \textbf{64.6}        & \textbf{65.2}        & \textbf{65.1}        & \textbf{60.6}        \\\cmidrule{1-17}
\multirow{6}{*}{\textbf{TD}} & \textit{w/o specialization}  &                      &                      &                      &                      &                      &                      &                      &                      &                      &                      &                      &                      &                      &                      &                      &       \multicolumn{1}{l}{}               \\
                                   & mBERT                        & 68.0                 & 64.3                 & 64.3                 & 56.1                 & 52.2                 & 53.4                 & 60.1                 & 55.3                 & 57.1                 & 52.0                 & 47.1                 & 49.0                 & 64.7                 & 67.1                 & 66.3                 & 58.5                 \\\cmidrule{2-17}
                                   & \textit{with specialization} &                      &                      &                      &                      &                      &                      &                      &                      &                      &                      &                      &                      &                      &                      &                      &       \multicolumn{1}{l}{}               \\
                                   & MLM                          & 69.0                 & 64.2                 & 65.2                 & \textbf{57.1}        & 52.6                 & 54.1                 & \textbf{61.5}        & 56.5                 & 58.7                 & \textbf{52.5}        & 47.2                 & 50.3                 & 65.1                 & 67.3                 & 67.3                 & 59.2                 \\
                                   & MTL-W (CLS)                  & \textbf{69.8}        & 64.4                 & \textbf{65.8}        & 56.9                 & \textbf{53.3}        & \textbf{54.5}        & 60.8                 & \textbf{57.6}        & \textbf{59.3}        & 51.1                 & 47.3                 & 50.3                 & 66.0                 & \textbf{68.1}        & 66.5                 & \textbf{59.4}        \\
                                   & MTL-W (CTX)                  & 69.2                 & \textbf{65.4}        & 64.9                 & 56.2                 & 53.2                 & 54.3                 & 59.3                 & 56.5                 & 59.3                 & 50.2                 & \textbf{48.0}        & \textbf{50.8}        & \textbf{66.4}        & 67.3                 & \textbf{67.6}        & 59.2\\ \bottomrule                
\end{tabular}%
}
\caption{Results of the extrinsic evaluation on \textit{age} data for age-specialized and non-specialized models for Sentiment Analysis (SA) and Topic Detection (TD). We report the F1-score.}
\vspace{-0.3em}
\label{tab:age_task}
\end{table*}

\subsection{Models and Baselines}
\label{subsec:baselines}
 In our first experiment, we employ mBERT~\citep{devlin-etal-2019-bert}, a multilingual PLM. We do so for efficiency reasons: the specialization procedure needs to be conducted only once on the multilingual intermediate training set and the resulting specialized model can then be applied to each of our country-specific fine-tuning portions. To validate our results and to control for the confounding factor of the multilingual representations, we later run additional experiments in which we resort to monolingual BERT variants for English, French, German and Danish.\footnote{We use the multilingual and monolingual PLM from Huggingface: \texttt{bert-base-multilingual-cased} (\textit{multilingual}),  \texttt{bert-base-cased} (\textit{English}), \texttt{bert-base-german-cased} (\textit{German}), \texttt{dbmdz/bert-base-french-europeana-cased} (\textit{French}) and \texttt{Maltehb/danish-bert-botxo} (\textit{Danish}).} As baselines, we report the performance of the non-specialized counterparts (i.e., we only fine-tune the PLMs on the training sets of the  downstream tasks) and compare them against our sociodemographic-specialized PLM variants, obtained after intermediate training on the mixed-gender or -age corpora (\S\ref{sec:method}): (1)~\textbf{MLM}: continual masked language modeling, (2)~\textbf{MTL-W (CLS)}: dynamic multi-task learning in which we couple the MLM objective and the sociodemographic class prediction on the \texttt{[CLS]} token representation, and (3)~\textbf{MTL-W (CTX)}: dynamic multi-task learning, in which we predict the sociodemographic class based on the representation of the averaged contextualized masked tokens. 

\vspace{-0.8em}
\subsection{Hyperparameters and Optimization}
During specialization, we fix the maximum sequence length to $128$ subword tokens. We train for $30$ epochs, in batches of $32$ instances and search for the optimal learning rate among the following values: $\{5\cdot 10^{-5}, 1\cdot 10^{-5}, 1\cdot 10^{-6}\}$. We apply early stopping based on development set performance (patience:~3~epochs). 
%
%
For downstream task evaluation, we train for maximum 20 epochs in batches of $32$ with the learning rate among the following values: $\{5\cdot 10^{-5}, 1\cdot 10^{-5}, 5\cdot 10^{-6}, 1\cdot 10^{-6}\}$. We also apply dev-set-based early stopping (patience: 5 epochs). In all experiments, we use Adam~\citep{kingma2014adam} as the optimization algorithm.

\section{Results and Discussion}
\label{sec:results}
We first discuss the results of the multilingual model (with and without specialization) across the five countries and the two sociodemographic factors for our three evaluation tasks~(\S\ref{subsec:initial_results}). We then thoroughly examine the effectiveness of the proposed sociodemographic specialization methods through a set of control experiments~(\S\ref{subsec:control_results}).
\vspace{-0.3em}
\setlength{\tabcolsep}{2.3pt}
\begin{table*}[t]
\centering
\footnotesize{
\begin{tabular}{llcccccccccccccccc}
\toprule
                                  &                              & \multicolumn{8}{c}{\textit{\textbf{gender}}}                                                                                                                                                   & \multicolumn{8}{c}{\textit{\textbf{age}}}                                                                                                                                                      \\
                                  \cmidrule(lr){3-10}\cmidrule(lr){11-18}
                                  &                              & \multicolumn{1}{c}{\textbf{AC-SA}}       & \multicolumn{1}{c}{\textbf{AC-TD}}       & \multicolumn{3}{c}{\textbf{SA}}                                    & \multicolumn{3}{c}{\textbf{TD}}                                    & \multicolumn{1}{c}{\textbf{AC-SA}}       & \multicolumn{1}{c}{\textbf{AC-TD}}       & \multicolumn{3}{c}{\textbf{SA}}                                    & \multicolumn{3}{c}{\textbf{TD}}                                    \\
                                  &   &     \multicolumn{1}{c}{$\Delta$}                 &       \multicolumn{1}{c}{$\Delta$}               &         \multicolumn{3}{c}{$\Delta$}                  &                    \multicolumn{3}{c}{$\Delta$}           &   \multicolumn{1}{c}{$\Delta$}                 &       \multicolumn{1}{c}{$\Delta$}               &         \multicolumn{3}{c}{$\Delta$}                  &                    \multicolumn{3}{c}{$\Delta$}  \\ \cmidrule(lr){3-3}\cmidrule(lr){4-4}\cmidrule(lr){5-7}\cmidrule(lr){8-10}\cmidrule(lr){11-11}\cmidrule(lr){12-12}\cmidrule(lr){13-15}\cmidrule(lr){16-18}
                                  & \textbf{Model}               & \textbf{}            & \textbf{}            & \multicolumn{1}{c}{\textbf{F}}           & \multicolumn{1}{c}{\textbf{M}}           & \multicolumn{1}{c}{\textbf{X}}           & \multicolumn{1}{c}{\textbf{F}}           & \multicolumn{1}{c}{\textbf{M}}           & \multicolumn{1}{c}{\textbf{X}}          & \textbf{}            & \textbf{}            & \multicolumn{1}{c}{\textbf{<35}}            & \multicolumn{1}{c}{\textbf{>45}}           & \multicolumn{1}{c}{\textbf{X}}           & \multicolumn{1}{c}{\textbf{<35}}            & \multicolumn{1}{c}{\textbf{>45}}            & \multicolumn{1}{c}{\textbf{X}}           \\ \midrule
\multirow{5}{*}{\textbf{US}} & \textit{w/o specialization}  & \multicolumn{1}{l}{} & \multicolumn{1}{l}{} & \multicolumn{1}{l}{} & \multicolumn{1}{l}{} & \multicolumn{1}{l}{} & \multicolumn{1}{l}{} & \multicolumn{1}{l}{} & \multicolumn{1}{l}{} & \multicolumn{1}{l}{} & \multicolumn{1}{l}{} & \multicolumn{1}{l}{} & \multicolumn{1}{l}{} & \multicolumn{1}{l}{} & \multicolumn{1}{l}{} & \multicolumn{1}{l}{} & \multicolumn{1}{l}{} \\                    
                                                     & BERT                         & 1.7                  & 0.5                  & 2.3                  & 2.6                  & 1.1                  & 1.4                  & 1.3                  & 0.8                  & 2.3                  & 2.3                  & 2.8                  & 0.8                  & -0.6                 & 0.8                  & 0.6                  & 2.8                  \\\cmidrule{2-18}
                                                     & \textit{with specialization} &                      &                      &                      &                      &                      &                      &                      &                      &                      &                      &                      &                      &                      &                      &                      &                      \\
                                                     & MLM                          & 1.3                  & -0.9                 & 1.1                  & 1.4                  & 0.9                  & 1.0                  & 0.7                  & 1.0                  & 1.6                  & 2.3                  & 0.5                  & 1.7                  & 2.1                  & 2.2                  & 1.5                  & 1.5                  \\
                                                     & MTL-W (CTX)                  & 2.5                  & 0.6                  & 0.9                  & 1.1                  & 0.6                  & 0.4                  & 0.4                  & 1.1                  & 6.0                  & 2.0                  & 1.2                  & 0.1                  & 1.7                  & 0.2                  & 0.3                  & 1.8                  \\
                                                     \midrule
\multirow{5}{*}{\textbf{Denmark}}                    & \textit{w/o specialization}  &                      &                      &                      &                      &                      &                      &                      &                      &                      &                      &                      &                      &                      &                      &                      &                      \\
                                                     & BERT                         & 2.1                  & 2.0                  & 3.1                  & 3.1                  & 3.2                  & 1.4                  & 1.5                  & 0.9                  & 10.6                 & 0.8                  & 4.5                  & 3.6                  & 3.1                  & 2.3                  & 2.2                  & 2.9                  \\\cmidrule{2-18}
                                                     & \textit{with specialization} &                      &                      &                      &                      &                      &                      &                      &                      &                      &                      &                      &                      &                      &                      &                      &                      \\
                                                     & MLM                          & 0.8                  & 0.7                  & 3.1                  & 2.5                  & 2.5                  & 1.0                  & 1.9                  & 1.3                  & 2.0                  & 2.3                  & 4.4                  & 5.3                  & 4.6                  & 2.1                  & 2.7                  & 3.5                  \\
                                                     & MTL-W (CTX)                  & 1.1                  & 0.6                  & 3.3                  & 2.8                  & 2.6                  & 1.3                  & 1.3                  & 1.8                  & 2.5                  & 2.0                  & 3.4                  & 2.8                  & 4.0                  & 2.8                  & 2.2                  & 2.5                  \\ \midrule
\multirow{5}{*}{\textbf{Germany}}                    & \textit{w/o specialization}  &                      &                      &                      &                      &                      &                      &                      &                      &                      &                      &                      &                      &                      &                      &                      &                      \\
                                                     & BERT                         & 0.3                  & 1.0                  & 0.4                  & 0.5                  & -0.2                 & 0.2                  & 0.5                  & 2.0                  & -0.1                 & 2.7                  & 1.0                  & 2.9                  & 2.1                  & 1.5                  & 2.1                  & 1.3                  \\\cmidrule{2-18}
                                                     & \textit{with specialization} &                      &                      &                      &                      &                      &                      &                      &                      &                      &                      &                      &                      &                      &                      &                      &                      \\
                                                     & MLM                          & 0.8                  & -0.4                 & 0.4                  & 0.5                  & -0.8                 & 0.0                  & -0.3                 & 0.6                  & -3.0                 & 3.1                  & 4.5                  & 2.7                  & 1.4                  & 0.7                  & 1.1                  & 1.2                  \\
                                                     & MTL-W (CTX)                  & -2.0                 & 2.0                  & 1.1                  & 1.3                  & -0.7                 & 0.2                  & -0.2                 & 1.1                  & 0.7                  & 3.3                  & 4.9                  & 1.6                  & -2.7                 & 4.2                  & 1.8                  & -0.1                 \\ \midrule
\multirow{5}{*}{\textbf{France}}  & \textit{w/o specialization}  &                      &                      &                      &                      &                      &                      &                      &                      &                      &                      &                      &                      &                      &                      &                      & \\
\multicolumn{1}{l}{}                                 & BERT                         & 0.1                  & 1.9                  & 1.2                  & 0.3                  & 0.8                  & 1.4                  & 2.1                  & 2.0                  & 0.2                  & 0.5                  & 0.2                  & 0.1                  & -1.2                 & -0.1                 & 2.1                  & 0.7                  \\\cmidrule{2-18}
\multicolumn{1}{l}{}                                 & \textit{with specialization} &                      &                      &                      &                      &                      &                      &                      &                      &                      &                      &                      &                      &                      &                      &                      &                      \\
\multicolumn{1}{l}{}                                 & MLM                          & 0.4                  & 1.1                  & 1.2                  & 0.6                  & -0.8                 & 0.4                  & 0.9                  & 1.3                  & -0.9                 & -0.2                 & 0.8                  & -0.1                 & 0.0                  & 1.3                  & 1.3                  & -0.1                 \\
\multicolumn{1}{l}{}                                 & MTL-W (CTX)                  & -0.7                 & 0.0                  & 1.6                  & 0.8                  & 0.6                  & 1.4                  & 0.5                  & 1.2                  & 1.4                  & 0.8                  & -0.7                 & 0.9                  & 1.8                  & 4.4                  & 3.4                  & -0.5                 \\ \midrule
\multirow{5}{*}{\textbf{UK}}      & \textit{w/o specialization}  &                      &                      &                      &                      &                      &                      &                      &                      &                      &                      &                      &                      &                      &                      &                      &                      \\
\multicolumn{1}{l}{}                                 & BERT                         & 1.3                  & 1.9                  & 2.4                  & 2.0                  & 2.5                  & 0.7                  & 1.2                  & 1.3                  & 0.6                  & 0.6                  & 1.4                  & 2.4                  & 1.8                  & 3.4                  & 1.0                  & 1.7                  \\\cmidrule{2-18}
\multicolumn{1}{l}{}                                 & \textit{with specialization} &                      &                      &                      &                      &                      &                      &                      &                      &                      &                      &                      &                      &                      &                      &                      &                      \\
\multicolumn{1}{l}{}                                 & MLM                          & 0.7                  & -0.5                 & 1.9                  & 0.6                  & 1.5                  & 0.6                  & 1.4                  & 0.2                  & 1.5                  & 0.4                  & 5.4                  & 5.3                  & 3.9                  & 3.7                  & 2.9                  & 2.7                  \\
\multicolumn{1}{l}{}                                 & MTL-W (CTX)                  & -0.2                 & -0.8                 & 0.8                  & 1.0                  & 0.2                  & 0.6                  & 0.9                  & 0.4                  & 2.8                  & 2.5                  & 3.0                  & 1.3                  & 2.1                  & 1.9                  & 2.3                  & 1.6                   \\
\bottomrule
\end{tabular}%
}
\caption{Results of the monolingual PLMs (BERT, germanBERT, danishBERT, frenchBERT) on Sentiment Analysis (SA), Topic Detection (TD), and Attribute Classification (AC-SA, AC-TD) with and without intermediate specialization for \emph{gender} and \emph{age} on the country-specific specialization portions of Trustpilot. We report the deltas in F1-score compared to the original results of the multilingual PLM (mBERT).} 
\vspace{-0.3em}
\label{tab:mono-multi_compare}
\end{table*}

\subsection{Overall Results}
\label{subsec:initial_results}
\vspace{-0.3em}
We report the intrinsic performance on predicting the sociodemographic groups (AC-SA/-TD,  Table~\ref{tab:gender_age_intrinsic}) as well as the extrinsic performance on Sentiment Analysis (SA) and Topic Detection (TD) on the group-specific test portions of the two sociodemographic factors (\textit{gender}: Table~\ref{tab:gender_task}; \textit{age}: Table~\ref{tab:age_task}). 

The rows for each task are segmented into two parts: at the top we show the baseline results, mBERT \textit{without any sociodemographic specialization}, and the following rows contain results for our proposed adaptation methods (\S\ref{sec:method}), continual mask language modeling (\textbf{MLM}), or dynamic multi-task learning (using the sequence representation token (\textbf{MTL-W (CLS)}) or the average masked token representation (\textbf{MTL-W (CTX)})).

\vspace{0.7em}
\noindent\textbf{Gender.} 
Across all specialization methods, we see gains on the intrinsic task of predicting the gender of an author (for both AC-SA and AC-TD portions, Table~\ref{tab:gender_age_intrinsic}) with average improvements of up to 1.8 percentage points compared to vanilla mBERT.
The highest gains come from our multi-task learning methods, MTL-W (CLS) and MTL-W (CTX), where the association between linguistic factors and sociodemographics is directly enforced. 

Also, the results for the extrinsic evaluation on SA and TD (Table~\ref{tab:gender_task}) hold promise for successful task-agnostic gender specialization: on average, across the 5 countries with gender-specific (F, M) and gender-agnostic (X) fine-tuning and testing portions, we observe gains for all specialization methods of up to 1.3 percentage points for SA and 0.9 percentage points for TD. 

\vspace{0.7em}
\noindent\textbf{Age.} 
We observe similar improvements when studying age: in the intrinsic evaluation, i.e., predicting the age group of an author, our specialized models perform on average up to 2.7 percentage points better than their non-specialized counterpart (Table~\ref{tab:gender_age_intrinsic}). Similarly, when fine-tuning and testing on the age-specific (<35,  >45) and mixed (X) fine-tuning portions, the specialized models excel with 1.3 percentage points improvement for SA and 0.9 percentage points improvement for TD (Table~\ref{tab:age_task}).  

Again, this points to the effectiveness of sociodemographic specialization through intermediate training. 
Surprisingly, the multi-task learning methods do not help for age classification (AC-SA and AC-TD age) but always lead to improvements for SA and TD tasks. Thus, the injected knowledge of age offers less discriminative power than knowledge of gender. Our intuition is that age affects the texts less than gender and is rather expressed in non-lexical ways (i.e., sentence structure). 

Overall, the results are in line with the findings from previous work~\citep{hovy-2015-demographic}: \textit{specialization for \textit{gender} and \textit{age} factors injects sociodemographic knowledge into  multilingual PLMs}. 
This knowledge consistently improves downstream task performance. This effect is especially pronounced in multi-task learning, which encourages the model to establish an explicit connection between the linguistic elements (i.e., a sequence or a token) and the sociodemographic factors. 
This approach is also efficient: we only need to refine the PLM once (i.e., the knowledge is shared across the different languages) before fine-tuning on specific tasks.
\vspace{-0.2em}
\subsection{Control Experiments}
\label{subsec:control_results}

To validate our results, we pose four additional questions: 
(1) Will we get similar gains when using monolingual PLMs? Monolingual PLMs do not suffer from the curse of multilinguality~\citep{conneau-etal-2020-unsupervised} -- thus, intuitively, they have less potential for further improvement. 
(2)~Will we still see gains when specializing our models on data that contains knowledge about the sociodemographic factors but no knowledge about our domain at hand, i.e., product reviews? 
(3)~Can we quantitatively assess the impact of sociodemographics, domain, language, and injection method on the results?
(4)~Do the internal representations of the PLMs reflect the sociodemographic specialization?  
\begin{figure*}[ht]
	\centering
    \includegraphics[trim={0.3cm 0.5cm 1.3cm 0.4cm}, clip, width=\textwidth]{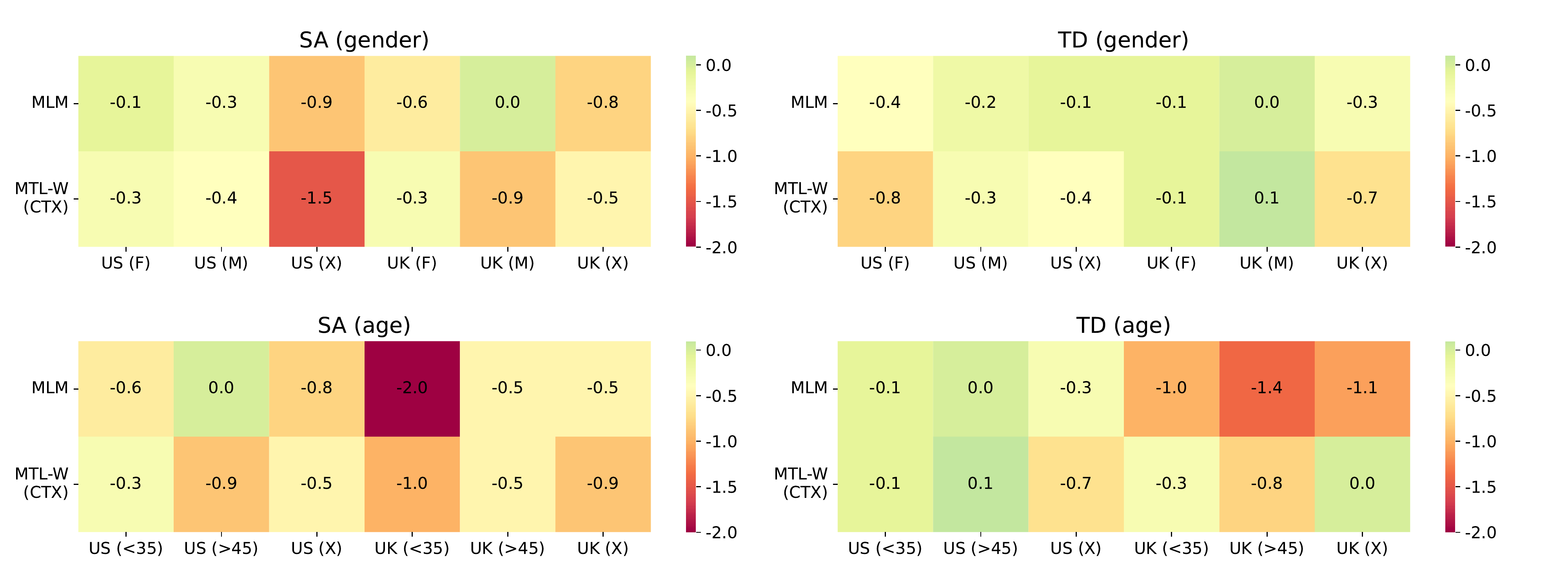}
    
    \vspace{-0.7em}
	\caption{Results on Trustpilot for Sentiment Analysis (SA) and Topic Detection (TD) when running the intermediate specialization on out-of-domain data (RtGender~\citep{voigt-etal-2018-rtgender} for \textit{gender} and BAC~\citep{schler2006effects} for \textit{age}). We report the delta in F1-score in comparison to the specialization on Trustpilot in-domain data.} 
	\label{fig:domain_heatmap}
	\vspace{-0.2em}
\end{figure*}

\setlength{\tabcolsep}{3.4pt}
\begin{table*}[h!]
\centering
\footnotesize{
\begin{tabular}{llgccccgcccc}
\toprule
 &  & \multicolumn{5}{c}{\textbf{RMSE}} & \multicolumn{5}{c}{\textbf{MAE}} \\\cmidrule(lr){3-7}\cmidrule(lr){8-12}
\textbf{Task} & \textbf{Selected features}& \textbf{in} & \textbf{ex-D} & \textbf{ex-M} & \textbf{ex-S} & \textbf{ex-C} & \textbf{in} & \textbf{ex-D} & \textbf{ex-M} & \textbf{ex-S} & \textbf{ex-C} \\\midrule
 \multicolumn{12}{l}{\textit{\textbf{gender}}} \\ \midrule
\textbf{AC-SA} & \begin{tabular}[c]{@{}l@{}}\textsc{Denmark} (4.5); \textsc{France} (3.5); \textsc{MLM} (0.9);\\ \textsc{MTL-W (CTX)} (0.9); \textsc{mono} (0.7) \end{tabular} \hspace{0.5em}& 0.58 & - & 0.66 & - & 1.61 & 0.58 & - & 0.52 & - & 1.27 \\\midrule
\textbf{AC-TD} & \begin{tabular}[c]{@{}l@{}}\textsc{UK} (5.5); \textsc{Denmark} (4.2); \textsc{MLM} (1.0);\\ \textsc{MTL-W (CTX)} (0.9); \textsc{mono} (0.5)\end{tabular} & 0.51 & - & 0.56 & - & 2.40 & 0.41 & - & 0.46 & - & 2.20 \\\midrule
\textbf{SA} &\begin{tabular}[c]{@{}l@{}}\textsc{UK} (5.9); \textsc{Denmark} (3.1);\\ \textsc{MLM} (0.8); \textsc{MTL-W (CTX)} (1.0);\\ \textsc{F} (1.3); \textsc{mono} (1.2); \textsc{in-domain} (0.5) \end{tabular}  & 1.07 & 1.08 & 1.20 & 1.38 & 2.27 & 0.68 & 0.71 & 0.82 & 1.07 & 1.90 \\\midrule
\textbf{TD} & \begin{tabular}[c]{@{}l@{}}\textsc{UK} (2.6); \textsc{US} (3.7);\\ \textsc{MLM} (0.5); \textsc{MTL-W (CTX)} (0.6);\\ \textsc{F} (1.2); \textsc{mono} (1.0); \textsc{in-domain} (0.5) \end{tabular} & 0.42 & 0.44 & 0.63 & 0.97 & 9.04 & 0.33 & 0.34 & 0.51 & 0.79 & 7.20 \\\midrule \midrule
 \multicolumn{12}{l}{\textit{\textbf{age}}} \\ \midrule
\textbf{AC-SA} & \begin{tabular}[c]{@{}l@{}}\textsc{Denmark} (7.5); UK (7.4); \textsc{MLM} (1.5);\\ \textsc{MTL-W (CTX)} (0.2); \textsc{mono} (1.8) \end{tabular} &1.66 & - & 1.89 & - & 4.37 & 1.13 & - & 1.29 & - & 4.03 \\\midrule
\textbf{AC-TD} & \begin{tabular}[c]{@{}l@{}}\textsc{UK} (5.1); \textsc{Denmark} (6.3); \textsc{MLM} (0.9);\\ \textsc{MTL-W (CLS)} (0.2); \textsc{mono} (1.6) \end{tabular}  & 0.79 & - & 1.11 & -  & 3.43 & 0.64 & - & 0.96 & - & 2.98 \\\midrule
\textbf{SA} &\begin{tabular}[c]{@{}l@{}}\textsc{UK} (9.7); \textsc{Denmark} (9.1);\\ \textsc{MLM} (0.9); \textsc{MTL-W (CTX)} (1.0); \\ \textsc{mono} (2.0); \textsc{in-domain} (1.0)\end{tabular}  & 1.15 & 1.18 & 1.48 & 1.16 & 3.46 & 0.91 & 0.96 & 1.21 & 0.91 & 3.85 \\\midrule
\textbf{TD} & \begin{tabular}[c]{@{}l@{}}\textsc{UK} (8.7); \textsc{US} (7.6);\\ \textsc{MLM} (0.5); \textsc{MTL-W (CLS)} (0.6); \\\textsc{mono} (1.9); \textsc{<35} (2.0); \textsc{in-domain} (0.8) \end{tabular} & 1.50 & 1.52 & 1.75 & 1.98 & 5.36 & 1.06 & 1.06 & 1.20 & 1.51 & 6.33\\
\bottomrule
\end{tabular}%
}
\vspace{-0.3em}
\caption{Results of our meta-regression analysis. We report the performance of predicting the results of our previous experiments on Sentiment Analysis (SA), Topic Detection (TD), and Attribute Classification (AC-SA and AC-TD) for \textit{gender} and \textit{age} when including all features (\textbf{in}) and excluding (\textbf{ex}) individual features: domain (\textbf{ex-D}), monolingual / multilingual (\textbf{ex-M}), sociodemographic factor (\textbf{ex-S}), country (\textbf{ex-C}). For each task, when including all features (column: \textbf{in}), we list the features with weights $\geq$ 0.2 (\textbf{selected features}) and provide their assigned weights in parenthesis. RMSE=Root Mean Square Error; MAE=Mean Absolute Error.}
\vspace{-0.3em}
\label{tab:meta}
\end{table*}

\vspace{0.7em}
\noindent\textbf{Monolingual vs Multilingual PLMs.}
We compare the results across five countries when leveraging monolingual vs. multilingual BERT~\citep{devlin-etal-2019-bert}, respectively (Table~\ref{tab:mono-multi_compare}). 
Unsurprisingly, the monolingual PLMs outperform their multilingual counterpart on all three tasks in most cases with gains of up to 2.5 percentage points for AC-SA gender. 
Further, we note that the performance improvement from before, which we attribute to our sociodemographic specialization, vanishes (i.e., < +0.5\% across the board). This finding challenges our previous hypothesis: are the performance gains we observed for the multilingual PLM obtained from injecting the sociodemographic knowledge?

\vspace{0.7em}
\noindent\textbf{In-domain vs Out-of-domain.}
To further investigate to what extent domain knowledge (i.e., knowledge about product reviews) is driving the performance improvements, we conduct additional experiments in which we control for this factor. Concretely, we select two out-of-domain corpora, which should still contain the sociodemographic knowledge we seek to specialize for: RtGender~\citep{voigt-etal-2018-rtgender} for \textit{gender} and Blog Authorship Corpus~\citep[BAC;][]{schler2006effects} for \textit{age}. RtGender consists of texts from online social media platforms\footnote{Rtgender is collected from five socially and topically diverse sources: Facebook (Politicians), Facebook (Public Figures), TED, Fitocracy, and Reddit.} where the author profiles indicate the \textit{gender}.  Similarly, BAC consists of texts from a blogging website (blogger.com). In case we still see improvements after specializing the models on these out-of-domain corpora, the gains should solely stem from the sociodemographic knowledge -- and thus prove the injection effective. To test this, we sample 100K and 50K instances each from the corpora and then specialize BERT via MLM and MTL-W (CTX). Since texts from both corpora are in English and to eliminate the confounding multilingual dimension, we evaluate on the Trustpilot fine-tuning data from English-speaking countries (UK and US) only and use the English monolingual PLM (original BERT).  
As expected, the out-of-domain specialization deteriorates the downstream task performance in most cases (Figure~\ref{fig:domain_heatmap}). Especially explicitly coupling linguistic aspects with the sociodemographics through MTL-W (CTX) seems to hurt the performance. 
The observation once again leads us to question whether the performance gains we observe in the multilingual scenario can be attributed to sociodemographic adaptation.
\begin{figure*}[t!]
\centering
\begin{subfigure}{\textwidth}
  \includegraphics[trim={1.3cm 1.6cm 0cm 1.7cm},clip,width=\textwidth, ]{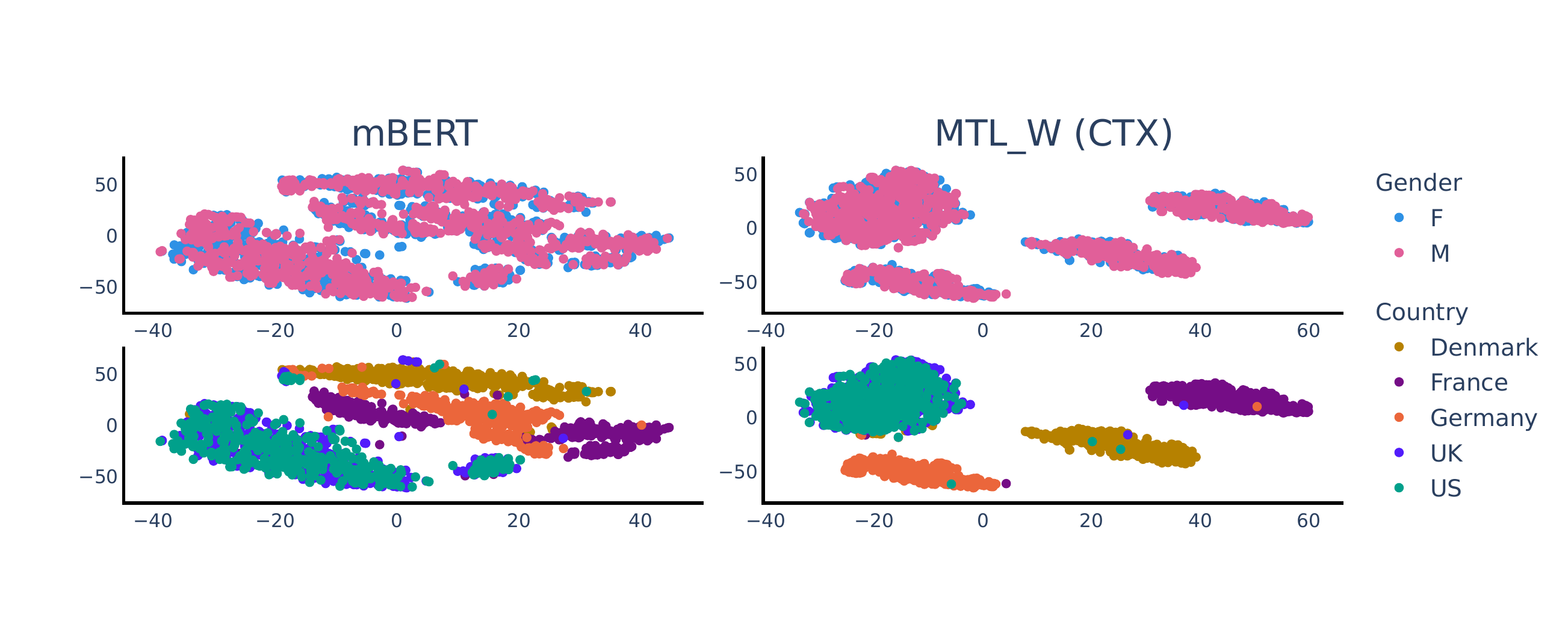}
  \caption{Gender}
\end{subfigure} 
\begin{subfigure}{\textwidth}
  \includegraphics[trim={1.3cm 1.6cm 0cm 1.7cm},clip,width=\textwidth]{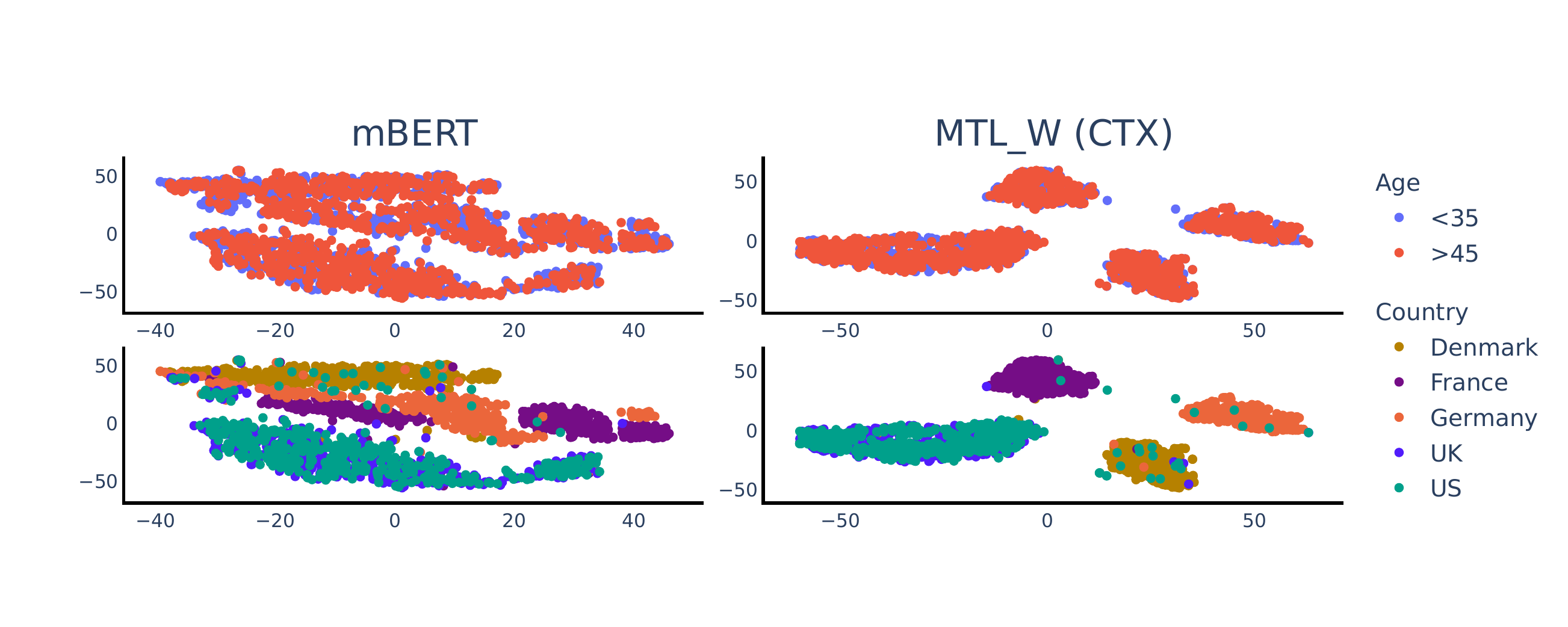}
  \caption{Age}
\end{subfigure}
\vspace{-0.5em}

\caption{Results of our multilingual qualitative analysis. We show a tSNE visualization of review texts embedded with a non-specialized (mBERT) and  specialized (MTL-W (CTX)) model. We plot 2K instances for (a) \emph{gender} and (b) \emph{age}. Colors indicate the sociodemographic subgroup (upper figures) and countries (lower figures), respectively.}
\vspace{-0.7em}
\label{fig:mbert_ctx_test_all_resize}
\end{figure*}

\vspace{0.7em}
\noindent\textbf{Meta-regression Analysis.}
Next, we quantitatively assess the impact of each of the factors involved in our study: the country of the fine-tuning texts (e.g., France), the method used (e.g., MLM), whether we run the specialization in-domain or out-of-domain, whether we use a monolingual or multilingual PLM, and the sociodemographic group the fine-tuning texts belong to (e.g., F, M, or X). To do so, we treat these factors as individual features and fit a linear regression to predict the final F1-scores obtained for each task (in all previously discussed Tables).  
We study the feature weights when including all features (\textbf{in}), and also run an ablation study to test how the performance scores (Root Mean Squared Error (RSME) and Mean Absolute Error (MAE))) change when excluding certain feature groups~(\textbf{ex-}). 
We summarize our meta-analysis in Table~\ref{tab:meta}. For each task, we list the selected features (weights for \textbf{in} in parenthesis) paired with the RMSE and MAE scores.  
When including all features (\textbf{in}), the individual countries which the fine-tuning portions originate from receive the highest weights across the board. Similarly, when removing the country information (\textbf{ex-C}), errors drastically increase (up to 8.62 points RSME increase). These two observations point to the generally high importance of the fine-tuning language and confirm our findings from the monolingual experiments. 
In contrast, adding or removing information of the sociodemographic group which the fine-tuning portion belongs to has a moderate impact only. Specifically for SA, adding or removing \textit{age} factors (\textbf{ex-S} in \emph{age}) has \emph{no} influence with respect to the task performance. We note similarly moderate changes when excluding information about the domain (\textbf{ex-D}).
Overall, we conclude that language seems to play the dominant role for the specialization results.

\begin{figure*}[t!]
\centering
\begin{subfigure}{0.48\textwidth}
  \includegraphics[trim={1.3cm 1.6cm 0cm 1.7cm},clip,width=\textwidth]{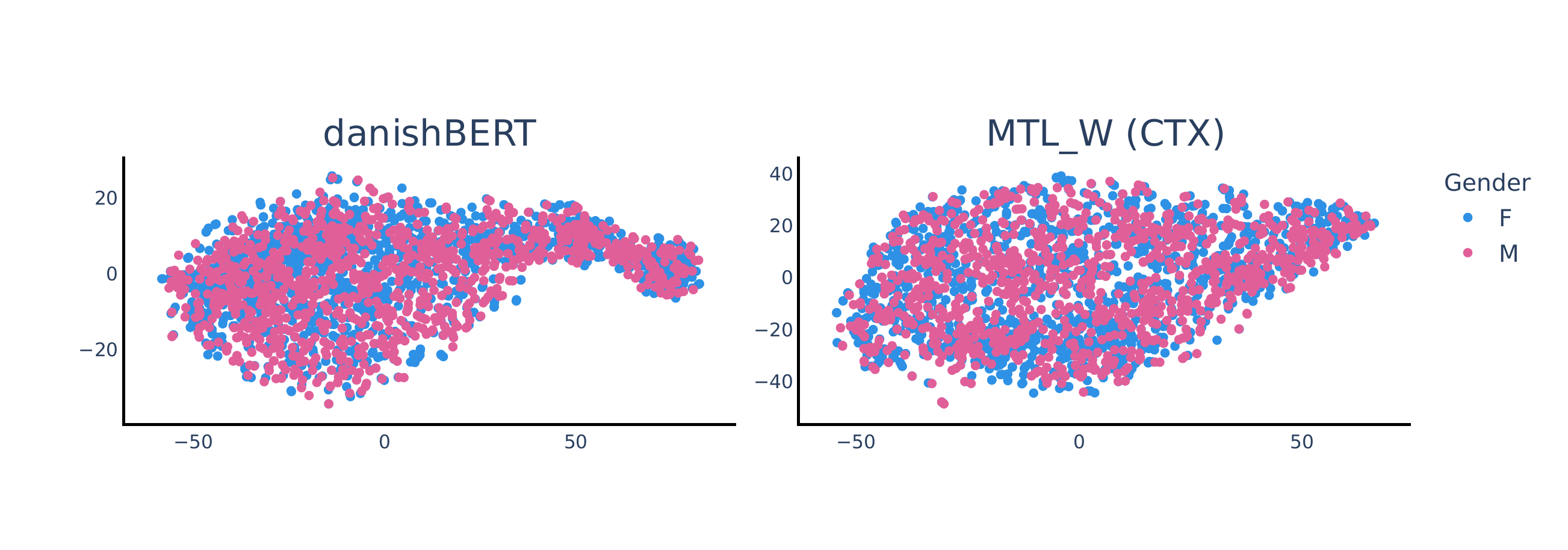}
  \caption{Denmark (gender)}
\end{subfigure}
\quad
\begin{subfigure}{0.48\textwidth}
  \includegraphics[trim={1.2cm 1.6cm 0cm 1.7cm},clip,width=\textwidth]{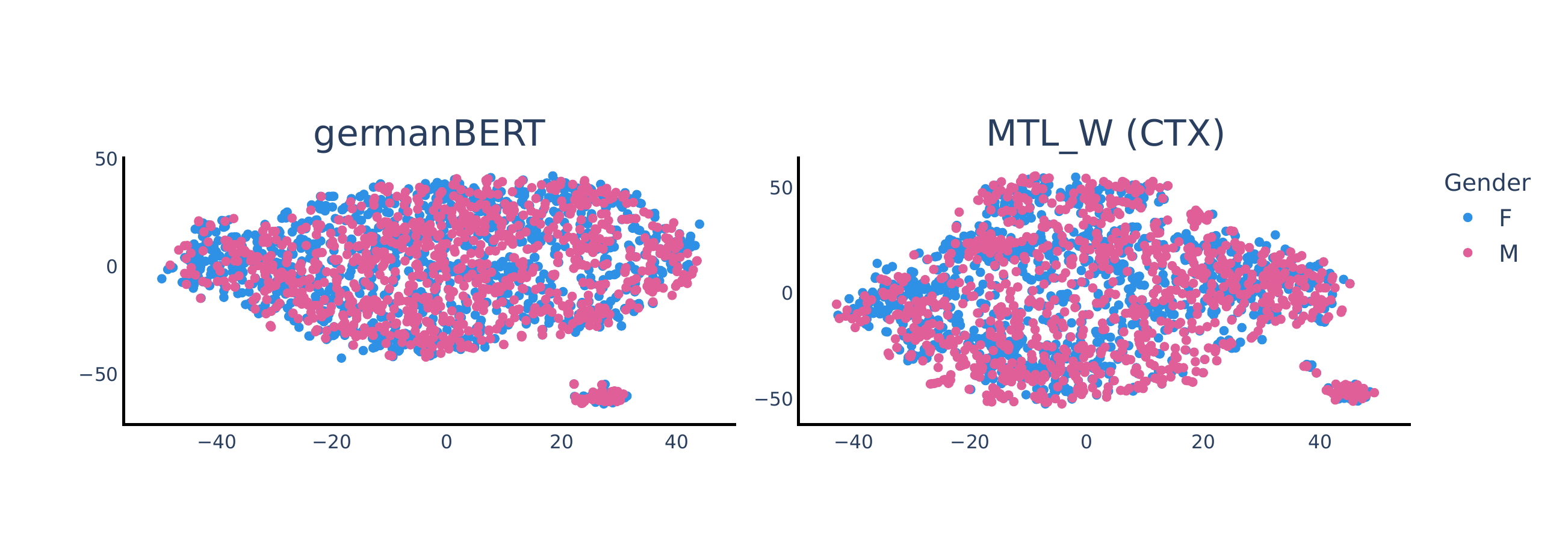}
  \caption{Germany (gender)}
\end{subfigure}
\begin{subfigure}{0.48\textwidth}
  \includegraphics[trim={1.3cm 1.6cm 0cm 1.7cm},clip,width=\textwidth]{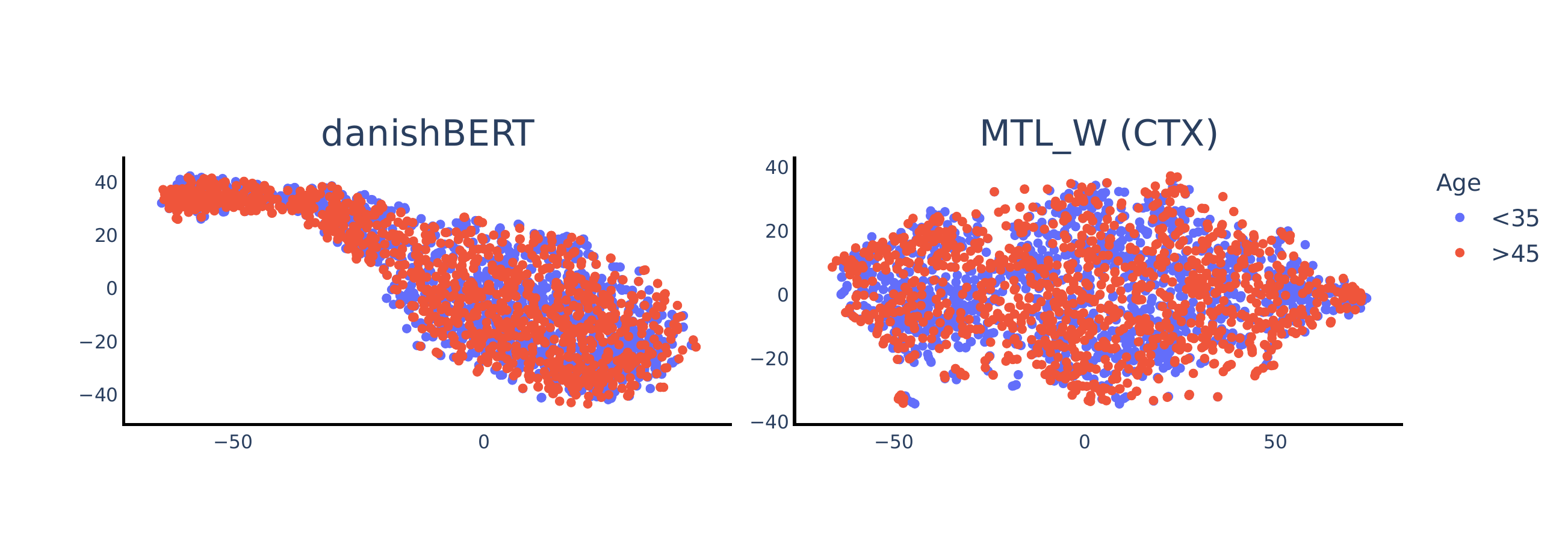}
  \caption{Denmark (age)}
\end{subfigure}
\quad
\begin{subfigure}{0.48\textwidth}
  \includegraphics[trim={1.2cm 1.6cm 0cm 1.7cm},clip,width=\textwidth]{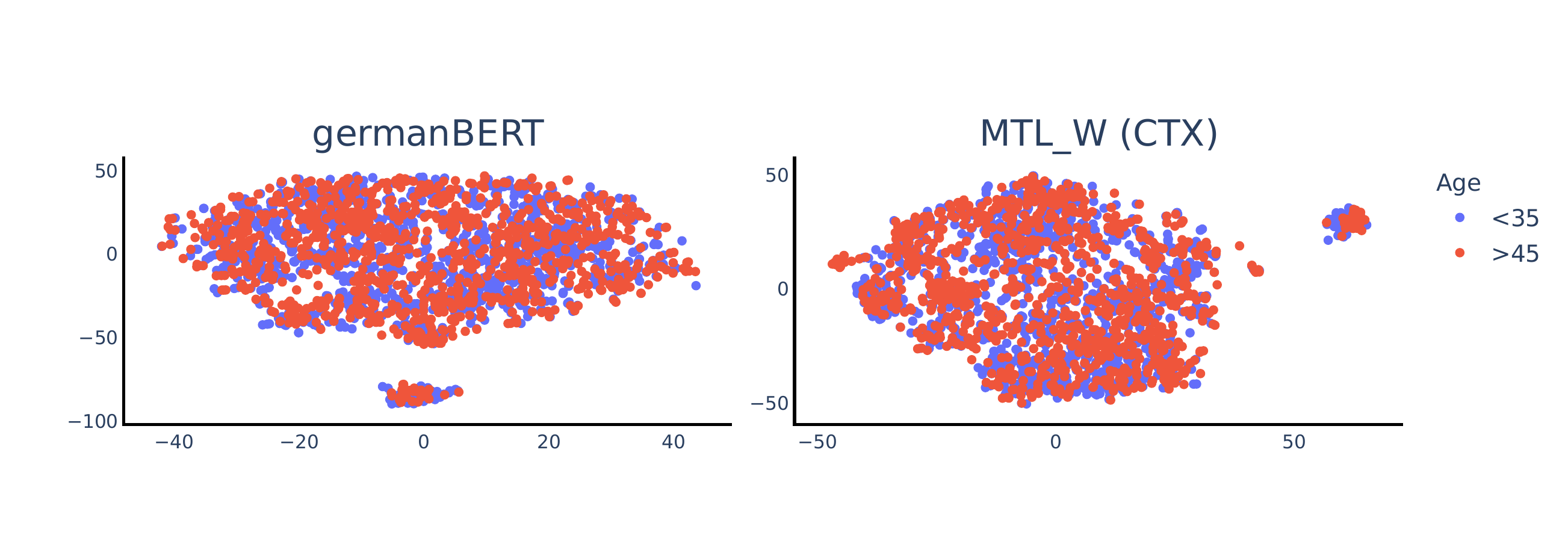}
  \caption{Germany (age)}
\end{subfigure}
\vspace{-0.7em}
\caption{Results of our monolingual qualitative analysis. We show a tSNE visualization of texts embedded with non-specialized (danishBERT, germanBERT) and specialized (MTL-W (CTX)) monolingual PLMs. We plot 2K instances for two countries (Denmark, Germany) and two sociodemographic factors (\textit{age}, \textit{gender}).}
\vspace{-0.7em}
\label{fig:monobert_ctx_test}
\end{figure*}

\vspace{0.7em}
\noindent\textbf{Qualitative Analysis.}
Finally, we visualize the \texttt{[CLS]} representations obtained when feeding the reviews through the encoder of the monolingual and multilingual BERTs before and after sociodemographic specialization with MTL-W (CTX). We randomly sample 2K instances from the development set of the specialization corpora.\footnote{Note that the dev set was originally kept for evaluating MLM training performance. The instances are equally distributed over 5 countries and 2 sub-groups of each sociodemographic factor.} 
For plotting, we reduce the dimension with the t-distributed stochastic neighbor embedding~\citep[tSNE;][]{JMLR:v9:vandermaaten08a} and color the points representing texts according to the sociodemographic group of the author. 
As illustrated in Figure~\ref{fig:mbert_ctx_test_all_resize}, representations obtained from the \textbf{MTL-W (CTX)}-specialized model arrange more clearly in clusters than the original mBERT. However, sociodemographic groups in \textit{gender} and \textit{age} do not cluster at all. In contrast, when coloring according to country, clear boundaries emerge.  
To further disentangle the situation, we also visualize the embeddings of the monolingual  variants in Figure~\ref{fig:monobert_ctx_test}. Although \textbf{MTL-W (CTX)} shows more assembly than the original BERTs, there is no significant clustering phenomenon for sociodemographic subgroups. We conclude: although our sociodemographic adaptation methods help improve the downstream performance, the gains do not stem from the sociodemographic knowledge itself, but more from distinguishing the diversified languages and other associated effects needed for the downstream tasks. 

\vspace{-0.8em}




\section{Related Work}


\paragraph{Intermediate Training (Adaptation).}
\vspace{-0.2em}
Intermediate language modeling on texts from the same or similar distribution as the downstream data has been shown to lead to improvements on various NLP tasks~\citep[e.g.,][]{gururangan-etal-2020-dont}. During this process, the goal is to inject additional information into the PLM and thus specialize the model for a particular domain~\citep[e.g.,][]{aharoni-goldberg-2020-unsupervised,hung-etal-2022-ds} or language~\citep[e.g.,][]{glavas-etal-2020-xhate} or to encode other types of knowledge such as common sense knowledge~\citep[e.g.,][]{lauscher-etal-2020-common}, argumentation knowledge~\citep[e.g.,][]{holtermann-etal-2022-fair}, or geographic knowledge~\citep[e.g.,][]{hofmann2022geographic}. 

For instance, \citet{glavas-etal-2020-xhate} and \citet{hung-etal-2022-multi2woz} perform language adaptation through intermediate MLM in the target languages with filtered text corpora, demonstrating substantial gains in downstream zero-shot cross-lingual transfer for abusive language detection and for dialog tasks, respectively. \citet{hung-etal-2022-ds} propose a  computationally efficient approach by employing domain-specific adapter modules. They show that their domain adaptation approach leads to improvements in task-oriented dialog. These specialization approaches mostly rely on a single objective (e.g., masked language modeling on ``plain'' text  data). Instead, \citet{hofmann2022geographic} conduct geoadaptation by coupling MLM with a token-level geolocation prediction in a dynamic multi-task learning setup. In this work, we adopt a similar approach. 

\vspace{-0.3em}
\paragraph{Sociodemographic Specialization.}
Language preferences vary with user demographics~\citep{loveys-etal-2018-cross}, and accordingly, several studies have leveraged sociodemographic information (e.g., gender, age, education) to obtain better language representations for various NLP tasks~\citep{volkova-etal-2013-exploring, garimella-etal-2017-demographic}. Recent research studies on sociodemographic adaptation mainly focus on (1) learning sociodemographic-aware word embeddings and do not work with large PLMs~\citep{hovy-2015-demographic} or (2) leveraging demographic information with special PLM architectures specifically designed for certain downstream tasks (e.g., empathy prediction~\citep{guda-etal-2021-empathbert}). The latter, however, do not consider a task-\textit{agnostic} approach to injecting sociodemographic knowledge into language models, and also focus on a monolingual setup only. Further, what roles the different factors (i.e., domain, language, sociodemographic aspect) in the specialization really play remains unexplored.
\vspace{-0.3em}

\section{Reproducibility}
\vspace{-0.2em}
To ensure full reproducibility of our results and fuel further research on sociodemographic adaptation in NLP, we release our code and data, which make our approach completely transparent: 
\url{https://github.com/umanlp/SocioAdapt}.
\vspace{-0.3em}

\section{Conclusion}
\vspace{-0.2em}
In this work, we thoroughly examined the effects of sociodemographic specialization of Transformers via straight-forward injection methods that have been proven effective for other types of knowledge. Initial results on extrinsic and intrinsic evaluation tasks  
using a multilingual PLM indicated the usefulness of our approach. However, running a series of additional experiments in which we controlled for potentially confounding factors (language and domain) as well as a meta-analysis indicate that the sociodemographic aspects only have a negligible impact on the downstream  performance. This observation is supported by an additional qualitative analysis. 
Overall, our findings point to the difficulty of injecting sociodemographic knowledge into Transformers and warrant future research on the topic for truly human-centered NLP. 

\section*{Limitations}
 In this work, we have focused on the sociodemographic adaptation of PLMs. We conducted our experiments with review texts covering five countries and four languages~\citep{hovy-2015-demographic}. It is worth pointing out that this data does not reflect the wide spectrum of (gender) identities~\citep{dev-etal-2021-harms, lauscher2022welcome} and is also limited with respect to its cultural variety~\citep{joshi-etal-2020-state}. Furthermore, while the criteria for selecting the data lead to balanced groups and potentially confounding factors have been monitored during the sampling process, potential harms might arise from unfair stereotypical biases in the data~\citep{blodgett2020}. 
 We hope that future research builds on top of our findings and explores other sociodemographic factors, other groups within these factors, and also other languages and countries. 

\section*{Acknowledgements}
The work of Anne Lauscher and Dirk Hovy is funded by the European Research Council (ERC) under the European Union’s Horizon 2020 research and innovation program (grant agreement No. 949944, INTEGRATOR). 

\bibliography{anthology,custom}
\bibliographystyle{acl_natbib_emnlp}



\end{document}